\documentclass{article}

\usepackage{arxiv}

\usepackage[utf8]{inputenc} % allow utf-8 input
\usepackage[T1]{fontenc}    % use 8-bit T1 fonts
\usepackage{hyperref}       % hyperlinks
\usepackage{url}            % simple URL typesetting
\usepackage{booktabs}       % professional-quality tables
\usepackage{amsfonts}       % blackboard math symbols
\usepackage{nicefrac}       % compact symbols for 1/2, etc.
\usepackage{microtype}      % microtypography
\usepackage{lipsum}

\usepackage{balance} % for balancing columns on the final page
\usepackage{amsmath}
\usepackage{mathrsfs}
\usepackage{bm}
\usepackage{booktabs}
\usepackage{amsfonts,amssymb}
\usepackage{colortbl}
\usepackage{color}
\usepackage{tikz}
\usepackage{multicol}
\usepackage{threeparttable}
\usepackage{makecell}
\usepackage{multirow}

\usepackage{algorithmicx}
\usepackage{algpseudocode}% http://ctan.org/pkg/algorithmicx
\usepackage[]{algorithm2e}
\usepackage{siunitx}
\usepackage{gensymb}

\usepackage{graphicx}

\usepackage{upgreek}
\usepackage{amsmath,amssymb,amsfonts}

\usepackage{amsmath}

\title{Evolving Multi-Objective Neural Network Controllers for Robot Swarms}

\author{
  Karl Mason\\
  School of Computer Science\\ 
  University of Galway\\
  Galway, Ireland\\
  \texttt{karl.mason@universityofgalway.ie} \\
   \And
 Sabine Hauert \\
  Bristol Robotics Laboratory\\ 
  University of Bristol\\ 
  Bristol, UK\\
  \texttt{sabine.hauert@bristol.ac.uk} \\
}

\begin{document}
\maketitle 

\begin{abstract}
Many swarm robotics tasks consist of multiple conflicting objectives. This research proposes a multi-objective evolutionary neural network approach to developing controllers for swarms of robots. The swarm robot controllers are trained in a low-fidelity Python simulator and then tested in a high-fidelity simulated environment using Webots. Simulations are then conducted to test the scalability of the evolved multi-objective robot controllers to environments with a larger number of robots. The results presented demonstrate that the proposed approach can effectively control each of the robots. The robot swarm exhibits different behaviours as the weighting for each objective is adjusted. The results also confirm that multi-objective neural network controllers evolved in a low-fidelity simulator can be transferred to high-fidelity simulated environments and that the controllers can scale to environments with a larger number of robots without further retraining needed.
\end{abstract}

% keywords can be removed


% footnote
% Proc. of the Autonomous Robots and Multirobot Systems (ARMS) 2023, Basilico et al. (eds.), May 9-10, 2023, https:// u.cs.biu.ac.il/ ~agmon/ arms2023/ . 2023.

\section{Introduction}\label{sec:Intro}

%what is the problem?
\let\thefootnote\relax\footnotetext{\textit{Proc. of the Autonomous Robots and Multirobot Systems (ARMS) 2023, Basilico et al. (eds.), May 9-10, 2023, \url{https://u.cs.biu.ac.il/~agmon/arms2023/}. 2023.}}

Many robotics tasks consist of multiple objectives. For example, manufacturing robots must accomplish tasks quickly and accurately \cite{chen2003multi}. These are conflicting objectives. Similarly, multiple objectives are also present in swarm robotics tasks. Minimizing both movement time and energy costs is a primary example of this \cite{mai2019multi}.

Current approaches to addressing these multi-objective robotic tasks often involve applying a multi-objective optimisation algorithm to optimise the actions of the robot for different objectives, e.g. for path planning \cite{nazarahari2019multi,sathiya2019evolutionary,wang2018car}. These approaches work well for determining the optimum paths under multiple objectives for a specific environment configuration. If the environment changes, e.g. if more robots are added to the environment, these optimisation based techniques must be reapplied to determine the new set of optimal paths under the new conditions. This requires additional computational time.

An alternative solution to this multi-objective robotics problem is presented in this paper using Neuroevolution, or evolutionary neural networks \cite{stanley2019designing}. Neuroevolution involves applying evolutionary algorithms to train the parameters of neural networks to solve a machine learning task, e.g. play Atari games \cite{hausknecht2014neuroevolution}, energy forecasting \cite{mason2018forecasting}, CPU utilization prediction \cite{mason2018predicting} and economic dispatch \cite{mason2018multi}. A multi-objective evolutionary neural network approach is proposed for developing swarm robot controllers. This proposed approach to developing multi-objective swarm robot controllers has the benefit of evolving a control policy that can still effectively control the robots even if the environmental conditions change and will not require the computational cost associated with retraining the robot controller.

The research presented in this paper makes the following contributions:

\begin{enumerate}
    \item To propose a Natural Evolution Strategies based evolutionary multi-object neural network approach for swarm robotics.
    
    \item To investigate the transferability of multi-objective neural network controllers trained in a low-fidelity simulator to a high-fidelity simulator.
    
    \item To determine if controllers trained in an environment with a small number of robots, can scale to larger robot swarms.
    
\end{enumerate}

The structure of the paper is as follows. Sections 2 will give an overview of the background literature relating to this research. Section 3 will outline the experimental methods, including simulator design, application of proposed approach and experiments conducted. Section 4 will present the results of the simulations conducted. Finally, Section 6 will present a conclusion for the research described in the paper.

\section{Background}

\subsection{Swarm Robotics}

Robot swarms typically have characteristics including: acting autonomously within their environment, homogeneity across robots, and utilizing local information only \cite{csahin2004swarm}. Robot swarms have many practical applications \cite{schranz2020swarm}, e.g. search and rescue \cite{bakhshipour2017swarm} and warehouse operations \cite{liu2017novel}.

There have been a number of publications that have explored the application of machine learning to robot swarms, e.g. in 2020 Tolstaya et al. applied a graph neural network to learn to control robots in a swarm \cite{tolstaya2020learning}. A 2021 paper by Dorigo et al. mentions the application of machine learning to robot swarms as one of the future directions of research in swarm robotics \cite{dorigo2021swarm}.

Research published recently in the literature has recognised that the task of robot path planning often consists of multiple objectives \cite{ma2018multi}, e.g. to minimize the time taken to locate a target, to minimise chance of collisions, to maximize energy efficiency, etc. Each of these objectives are important and should be considered when determining the behaviour of the robot. This motivates the research presented in this paper to develop robot controllers that can take preferences for each objective as input to influence the behaviour of robot swarms.

Multiple studies have been published in the literature that apply evolutionary methods to swarm robotics. Birattari et al. provide a manifesto for automatic off-line design in the area of robot swarms \cite{birattari2019automatic}. This paper discusses existing applications of evolutionary neural networks to robot controllers. Floreano et al. provide a comprehensive account of applications of evolutionary computing to robotics \cite{Floreano2008}. Evolutionary methods have been successfully applied in swarm robotics tasks. Hauert et al. applied evolutionary neural networks to control individuals in a swarm of simulated Micro Air Vehicles (MAVs) \cite{hauert2009evolved}. Evolutionary methods have also been applied to swarm robotics by evolving behaviour trees \cite{jones2019onboard,jones2018evolving}. Recent studies have applied MAP-Elites for swarm robotics \cite{kaiser2022minimize}. These studies demonstrate the effectiveness of evolutionary methods for swarm robotics tasks. The research outlined in the paper builds on these previous studies by extending evolutionary swarm robot controllers to multi-objective tasks using Natural Evolution Strategies.

There have been a number of studies published in the literature that explore multi-objective control in robot swarms. Mai et al. developed a Multi-Objective collective search strategy for robot swarms based on the Particle Swarm Optimisation algorithm \cite{mai2019multi}. Miao et al. proposed a Multi-objective region reaching controller for a swarm of robots \cite{miao2019multi}. These studies demonstrate the utility of multi-objective control and path planning for robot swarms. These studies do not consider the use of multi-objective neural network controllers for robot swarm control, as this research paper presents.

% **maybe more here***

\subsection{Evolutionary Neural Networks}

Neural networks are machine learning models that take inspiration from the brain. The field of Evolutionary Neural Networks (or Neuroevolution) utilize evolutionary algorithms and principals to train the parameters of neural networks \cite{galvan2021neuroevolution}. Target network outputs are not required when evolving neural networks, only a fitness function. Neuroevolution has been shown to be a competitive approach when compared to reinforcement learning algorithms \cite{salimans2017evolution,mason2021building}. 
% Evolutionary neural networks have been applied to a wide range of problems, e.g. energy forecasting \cite{mason2018forecasting}, CPU utilization prediction \cite{mason2018predicting} and economic dispatch \cite{mason2018multi}.

\begin{algorithm}[h]
\caption{xNES Algorithm}
\begin {algorithmic}
\State \textbf{Initialize algorithm}\\
\While{curGen < maxGen}{\\
    \For{i in $\lambda$}{
            \State Sample $z_i \leftarrow \mathcal{N}(0,1)$ 
            \State $x_i \leftarrow \mu + \sigma \textbf{B} z_i$}
    \State calculate utilities $u$ by sorting $\{(z_i, x_i)\}$ w.r.t $f(x_i)$
    \State $G_\delta \leftarrow \Sigma_{i=1}^{\lambda} u_i \cdot z_i$
    \State $G_M \leftarrow \Sigma_{i=1}^{\lambda} u_i \cdot (z_i z_{i}^T - \textbf{I})$
    \State $G_\sigma \leftarrow \frac{trace(G_M)}{d}$
    \State $G_B \leftarrow G_M -G_\sigma \cdot \textbf{I}$
    \State $\mu \leftarrow \mu + \eta_\mu \cdot \sigma \textbf{B} \cdot G_\delta$
    \State $\sigma \leftarrow \sigma \cdot exp (0.5 \eta_\sigma \cdot  G_\sigma)$
    \State $\textbf{B} \leftarrow \textbf{B} \cdot exp (0.5 \eta_\textbf{B} \cdot  G_\textbf{B})$
    \State $curGen++$}
\end{algorithmic}
\label{Alg:xNES}
\end{algorithm}

There are many Neuroevolution algorithms that evolve the weights and architecture of the neural network, e.g. NeuroEvolution of Augmenting Topologies (NEAT) \cite{stanley2002evolving} and hyperNEAT \cite{d2014hyperneat}. Many studies implement methods such as evolutionary strategies to evolve only the weights of the network \cite{chen2019restart,pourchot2018importance,mason2021building}. Covariance Matrix Adaption Evolutionary Strategy (CMA-ES) \cite{hansen2003reducing} and Natural Evolution Strategies (NES) \cite{wierstra2014natural} are two well known evolutionary strategies. This research will use a variant of NES called Exponential Natural Evolution Strategies (xNES) \cite{glasmachers2010exponential} when evolving MO-NNs. It was selected as it is an effective optimisation algorithm.

%\subsection{Exponential Natural Evolution Strategies}
Algorithm \ref{Alg:xNES} describes the xNES algorithm. The algorithm samples $\lambda$ normally distributed solutions $z_i$. These are used to calculate $\lambda$ solutions $x_i = \mu + \sigma \textbf{B} z_i$, based on the center of the search distribution $\mu$, the normalized covariance factor $\textbf{B} = \textbf{A}/\sigma$, and $\sigma$ is the scalar step size. When the algorithm begins, $\textbf{A}$ is initialized as the identity matrix and $\sqrt[d]{|det(\textbf{A})|}$, where $d$ is the number of dimensions.

After solutions are sampled, the gradients of the objective function are calculated with respect to $\delta, M, \sigma$ and $\textbf{B}$. Where M is a $d \times d$ exponential map used to represent the covariance matrix $\textbf{C}$, where $\textbf{C = AA}^{T}$ and $\delta$ is the change in $\textbf{C}$.

\section{Experimental Methods}
\label{sec:Exp}

\subsection{Evolving Multi-Objective Neural Networks}
This research consists of evolving neural networks for a multi-objective swarm robotics task. The pseudocode in Algorithm \ref{Alg:MO_NN} illustrates the MO-NN training process.

\begin{algorithm}[h]
\caption{Evolving MO-NNs using xNES}
\begin {algorithmic}
\State \textbf{Initialize problem domain}
\State \textbf{Initialize algorithm}\\
\While{curGen < maxGen}{\\
    \State Sample $\lambda$ solutions $x$ (Alg \ref{Alg:xNES})\\
    \For{i in $\lambda$}{
        \State Set parameters of NNs as $x$
        \State fitness of NN $= networkFit_i = 0$\\
        \For{$\Updelta w=0$ to $1$}{  
            \State $w_1 = 1 - \Updelta w$
            \State $w_2 = \Updelta w$
            \State Reset environment\\
            \For{timeStep $t=0$ to $maxTime$}{\\
                \For{robot $r=1$ to $numRobots$}{
                    \State Robot $r$ observes state $s_{r,t}$ of the environment, i.e. sensor readings
                    \State Pass $w_1$, $w_2$ and $s_{r,t}$ into NN and record NN output $o$
                    \State Act in environment from NN output $o$, i.e. rotate and move
                    \State $currentFit =$ fitness of action using objective function (Equation \ref{eqn:curFit})
                    \State $networkFit_i += currentFit$
                    }
            }
        }    
    }
    \State calculate utilities $u$ by sorting $\{(z_i, x_i)\}$ w.r.t $networkFit_i$
    \State update gradients $G$, $\mu$, $\sigma$, and $\textbf{B}$ (Alg \ref{Alg:xNES})
    \State $curGen++$
    }
\end{algorithmic}
\label{Alg:MO_NN}
\end{algorithm}

 In Algorithm \ref{Alg:MO_NN}, $w_i$ is the weighting of objective $i$ and $\Updelta w$ is the change in objective weight. Each network must be capable of producing different outputs for different objective weightings given as input to the network. Mason et al. proposed a Multi-Objective Neural Network training architecture in 2018, that will be utilized in this research \cite{mason2018multi}. The distinction should be noted between evolving multi-objective neural networks and applying multi-objective evolutionary algorithms to train neural networks. The former consists of evolving a \textit{single set of network parameters} which enables the network to provide different outputs as the problem objective weighting varies, while the latter consists of utilizing a MO-EA, e.g. NSGA-II \cite{deb2002fast}, to evolve \textit{multiple sets of network parameters} for different objective weightings. This research evolves a single set of MO-NN parameters that are assigned the NN used to control each robot. The implementation of the MO-NN as a robot controller will be discussed in Section \ref{sec:EvoMONN_Robot}.

% \begin{algorithm}[H]
% \caption{Evolving MO-NNs using xNES}
% \begin {algorithmic}
% \State \textbf{Initialize problem domain}
% \State \textbf{Initialize algorithm}\\
% \While{curGen < maxGen}{\\
%     \State Sample $\lambda$ solutions $x$ (Alg \ref{Alg:xNES})\\
%     \For{i in $\lambda$}{
%         \State Set parameters of NNs as $x$
%         \State fitness of NN $= networkFit_i = 0$\\
%         \For{$\Updelta w=0$ to $1$}{
%             \State $w_1 = 1 - \Updelta w$
%             \State $w_2 = \Updelta w$
%             \State Reset environment\\
%             \For{timeStep $t=0$ to $maxTime$}{\\
%                 \For{robot $r=1$ to $numRobots$}{
%                     \State Robot $r$ observes state $s_{r,t}$ of the environment, i.e. sensor readings
%                     \State Pass $w_1$, $w_2$ and $s_{r,t}$ into NN and record NN output $o$
%                     \State Act in environment from NN output $o$, i.e. rotate and move
%                     \State $currentFit =$ fitness of action using objective function (Equation \ref{eqn:curFit})
%                     \State $networkFit_i += currentFit$
%                     }
%             }
%         }    
%     }
%     \State calculate utilities $u$ by sorting $\{(z_i, x_i)\}$ w.r.t $networkFit_i$
%     \State update gradients $G$, $\mu$, $\sigma$, and $\textbf{B}$ (Alg \ref{Alg:xNES})
%     \State $curGen++$
%     }
% \end{algorithmic}
% \label{Alg:MO_NN}
% \end{algorithm}

\subsection{Simulator Design}
\label{sec:simDesign}

% \begin{figure*}[h]
% \centerline{\includegraphics[width=0.6\textwidth]{SimulatedEnv.png}}
% \caption{Simulated robot (a) and arena (b).}
% \label{fig:simEnv}
% \end{figure*}

In order to train the proposed multi-objective neural network (MO-NN) controllers, a low-fidelity swarm robot simulator was developed in Python. This was necessary as evolving neural networks for control tasks requires many trial evaluations. This would be too computationally expensive for high-fidelity simulators.

\begin{figure}[h]
\centerline{\includegraphics[width=0.6\textwidth]{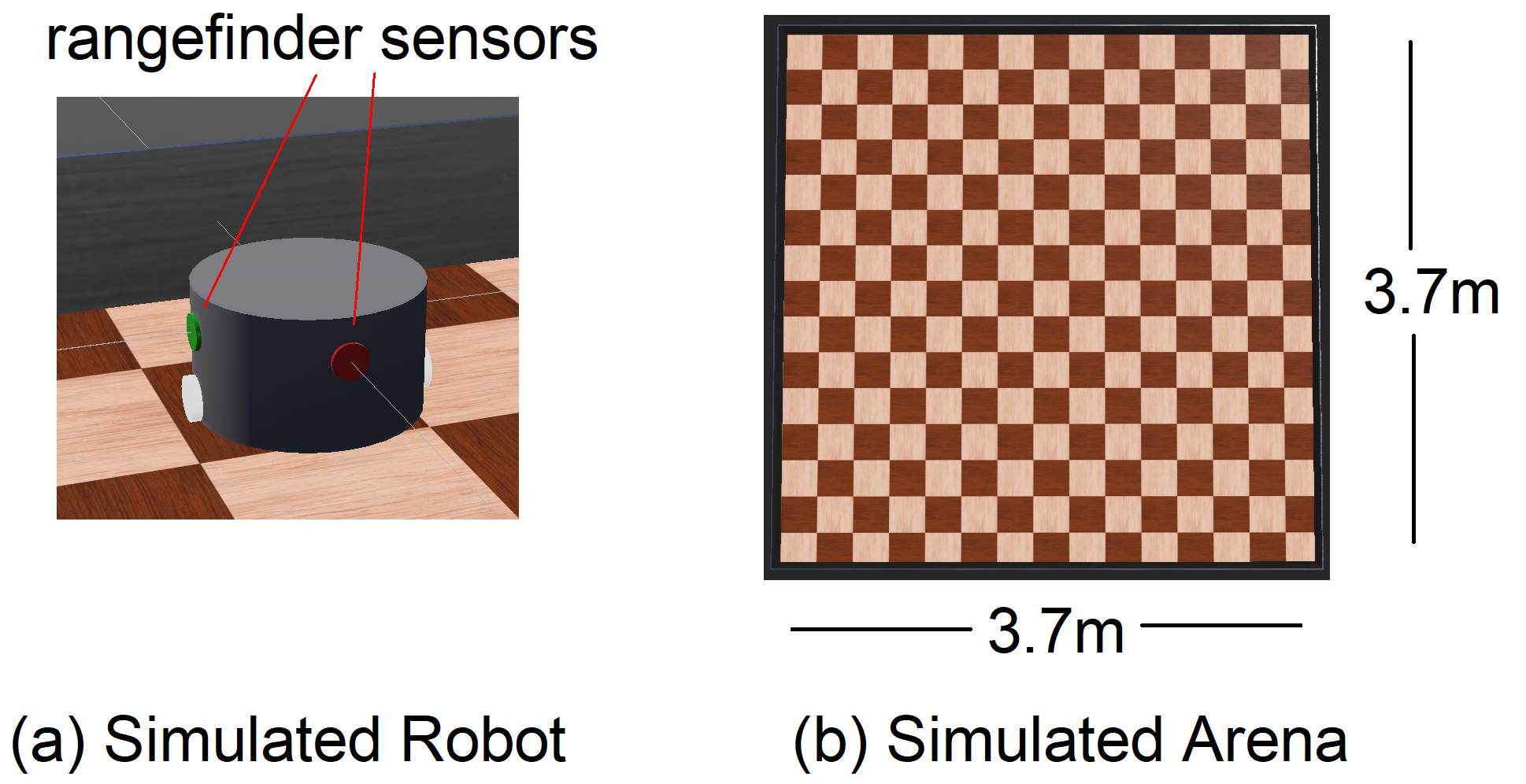}}
\caption{Simulated robot (a) and arena (b).}
\label{fig:simEnv}
\end{figure}

The low-fidelity simulator consisted of a $3.7m \times 3.7m$ arena with boundaries at the edge of the arena. Mobile robots were deployed within the arena. The simulated robots had a diameter of $0.25m$. Each robot had 4 rangefinder sensors, located at the front, back and on each side. Each robot can rotate $\pm 45\deg$ and can only move forward up to a maximum velocity of $2m/sec$. This simulator design is based on the DOTS swarm robot testbed \cite{jones2022dots}. 

A high-fidelity simulator was also developed using Webots \cite{Webots,Webots04}. This was used to test the evolved controllers. The parameters of the Webots simulator are the same as the Python simulator. Figure \ref{fig:simEnv} illustrates the simulated robot and arena.

Both simulators model non-holonomic robot drive. The low-fidelity simulator does not simulate any physics. Collision detection is implemented in the low-fidelity simulator to prevent a robot from colliding with another robot or the environment boundary. Physics is simulated in the high-fidelity Webots simulator. Collisions are not detected/prevented in the high-fidelity simulator.

\subsection{Multi-Objective Neural Network Robot Controllers}
\label{sec:EvoMONN_Robot}

\begin{figure*}[h]
\centerline{\includegraphics[width=0.85\textwidth]{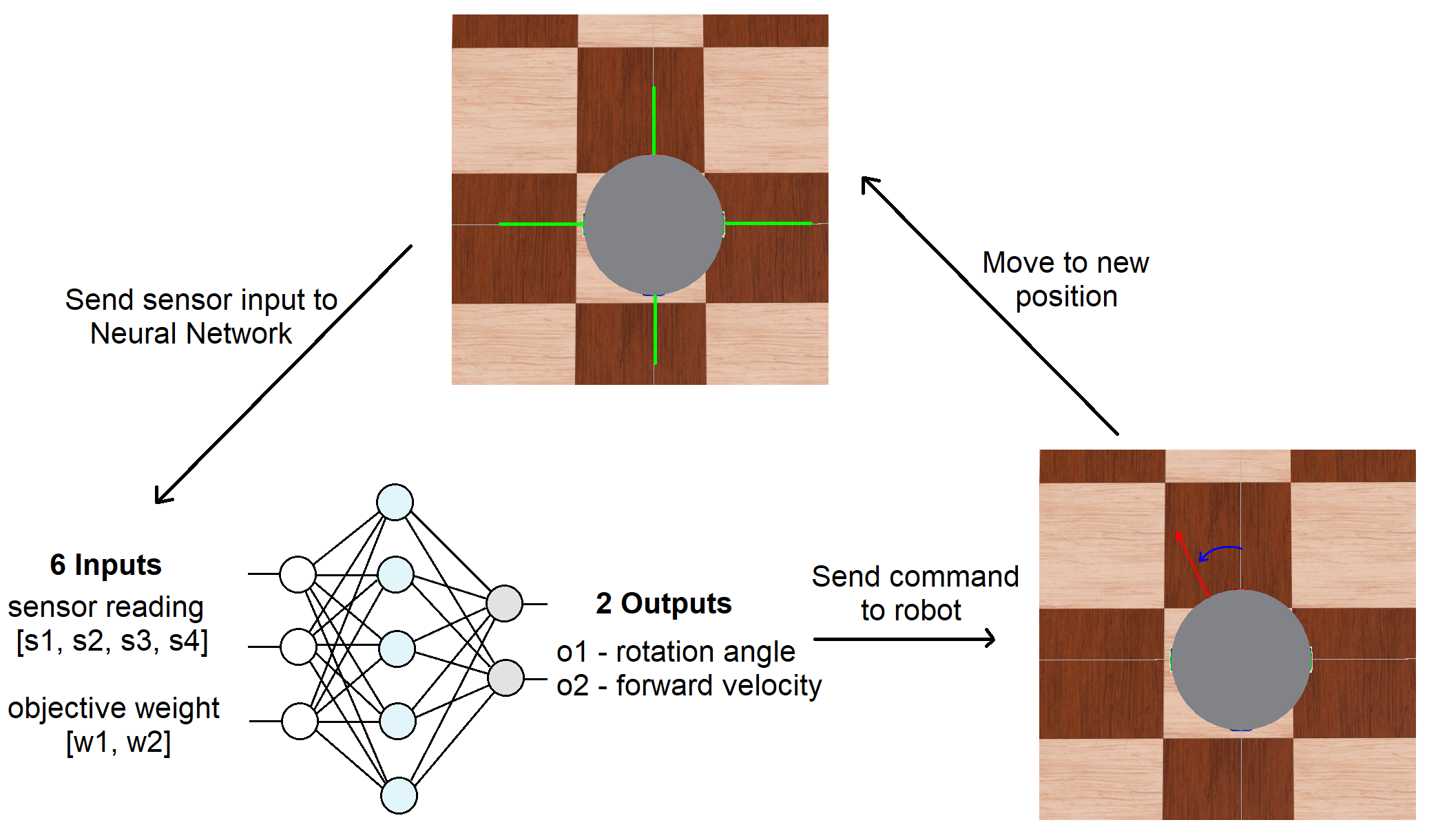}}
\caption{Neural network controller.}
\label{fig:NN_Control}
\end{figure*}

The implementation of the neural network controller as a robot controller is illustrated in Figure \ref{fig:NN_Control}. The robot senses its environment using 4 rangefinder sensors. These measure the distance between the sensor and the nearest object. These values are normalized and passed to the network as input. In addition to sensor measurements, two additional inputs are passed into the network representing the weightings assigned to each of the two objectives $[w_1, w_2]$. The network then does a forward pass and gives two outputs, i.e. robot commands. These commands are the rotation angle $[-45\deg, 45 \deg]$ and the forward velocity $[0m/sec, 2m/sec]$. The robot will move to a new position based on these commands.

\begin{figure*}[h]
\centerline{\includegraphics[width=0.75\textwidth]{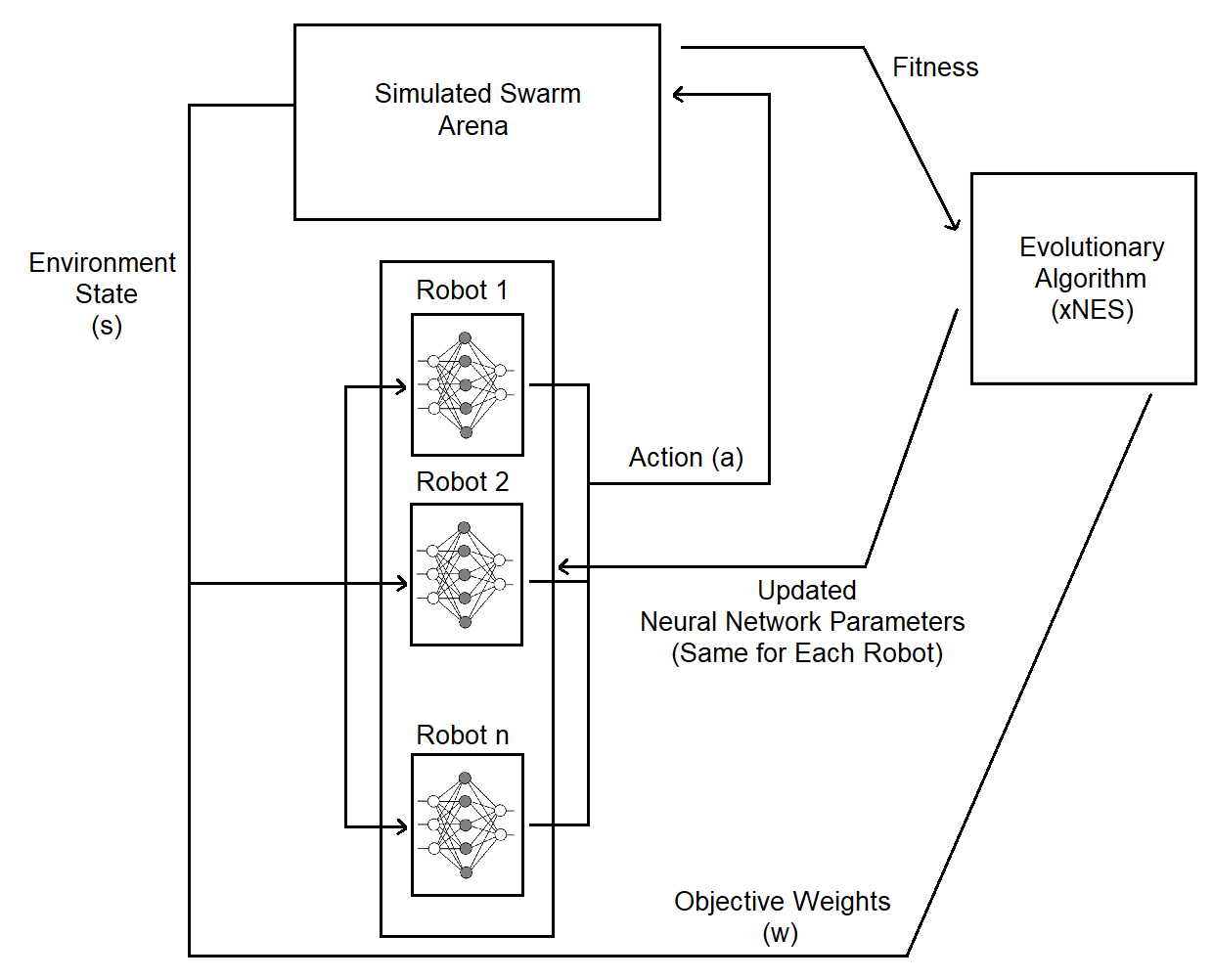}}
\caption{Evolving neural network swarm robot controller.}
\label{fig:EA_NN_Swarm}
\end{figure*}

Figure \ref{fig:EA_NN_Swarm} illustrates the training process of the MO-NNs for the robot swarm. Homogeneity is a key property of robot swarms, therefore the same network parameters are assigned to each NN controller for all robots in the swarm during each simulation.

The quality of each set of NN parameters $x_i$ is measured using the fitness function outlined in Equation \ref{eqn:totFit}.

\begin{equation}
  networkFit(NN(x_i)) = \sum\limits_{\Updelta w=0}^{1} \sum\limits_{t=1}^{t_{Max}} currentFit_{w,t}(NN(x_i))
  \label{eqn:totFit}
\end{equation}

\noindent where $\Updelta w$ is the change in objective weighting applied to $w_1$ and $w_2$. These are updated as $w_1 = 1 - \Updelta w$ and $w_2 = \Updelta w$ $\therefore$ $w_1 + w_2 = 1$  $\forall$  $\Updelta w$. The fitness at the current time-step for the current objective weights is calculated using Equation \ref{eqn:curFit}.

\begin{equation}
\begin{split}
  currentFit_{w,t}(NN(x_i)) = & w_1 \times - 1 \times Obj1_{w,t}(NN(x_i)) \\
  & + w_2 \times Obj2_{w,t}(NN(x_i)) 
\end{split}
\label{eqn:curFit}
\end{equation}

\noindent where $w_1$ is the weighting of objective 1 ($Obj1$) and $w_2$ is the weighting of objective 2 ($Obj2$). Objective 1 is to minimize the distance between robots and the origin (center of the arena), calculated using Equation \ref{eqn:obj1}. Objective 2 is to maximize the velocity of the robots, calculated using Equation \ref{eqn:obj2}. These two objectives were chosen as they are in direct conflict with one another. It should be noted here that $Obj1$ is multiplied by $-1$ as the overall optimisation problem is framed as a maximization problem. 

% swarm dist to origin obj
\begin{equation}
  Obj1_{w,t}(NN(x_i)) = \sum\limits_{r=1}^{numRobots} [|position_{x,r}|+ |position_{y,r}|]
  \label{eqn:obj1}
\end{equation}

\noindent where $position_{x,r}$ and $position_{y,r}$ represent the x and y positions of robot r, respectively. 

% swarm velocity obj
\begin{equation}
  Obj2_{w,t}(NN(x_i)) = \sum\limits_{r=1}^{numRobots} [|velocity_{x,r}|+ |velocity_{y,r}|]
  \label{eqn:obj2}
\end{equation}

\noindent where $velocity_{x,r}$ and $velocity_{y,r}$ represent the x and y velocities of robot r, respectively.

During training, each set of NN parameters $x_i$ is evaluated for 30 seconds of simulated time ($t_{Max} = 30sec$) for each increment of $\Updelta w$. The simulator time step is 1 second. The value of $\Updelta w$ is incremented between 0 and 1 in increments of 0.5. This is to ensure that each network is evaluated based on its ability to control the robot such that each objective is optimised with maximum weighting, i.e. when $w_1 = 1, w_2 = 0$ and $w_1 = 0, w_2 = 1$, and also its ability to control the robot such that each objective is weighted equally, i.e. when $w_1 = 0.5, w_2 = 0.5$. This number of $\Updelta w$ increments was selected to minimize training time. A smaller $\Updelta w$ increment can be implemented during training but would increase training time. Note, a smaller $\Updelta w$ increment can be applied when evaluating the MO-NN, irrespective of the increment size during training. 

% The results presented in Section \ref{sec:res_lowFid} utilize a $\Updelta w$ increment of $0.1$ when generating the Pareto Front, despite implementing a $\Updelta w$ increment of $0.5$ during training.

The NN architecture implemented in this research consisted of 6 input nodes (4 sensor inputs and 2 objective weights), 1 hidden layer with 5 nodes, and an outputs layer with 2 nodes (1 for rotation, 1 for forward movement). The network was evolved over 20,000 evaluations.

\subsection{Experimental Setup}

\begin{enumerate}
    \item \textbf{Evolving Neural Network Controller}.
This experiment consists of evolving the multi-objective neural network controller in the low-fidelity python robot simulator. The MO-NN is evolved in the low-fidelity python simulator with the Objective (obj) preferences: MO-NN 1: Obj 1 - Maximize velocity. Obj 2 - Minimize distance to origin (center of arena).

\item \textbf{Deployment to High-Fidelity Webots Simulator}
The next experiment is to test the performance of the best performing network evolved in the low-fidelity python simulator by deploying the network to control robots in a more realistic high-fidelity Webots simulator.

\item \textbf{Evaluating Evolved Controller on Larger Swarm Sizes}
The final experiment is to determine the scalability of the evolved MO-NN to a larger number of robots, specifically 5 and 10 robots. 

\end{enumerate}

\section{Results}

\subsection{Evolving Neural Network Controller in Low-Fidelity Python Simulator}
\label{sec:res_lowFid}

\begin{figure*}[h]
\centerline{\includegraphics[width=0.99\textwidth]{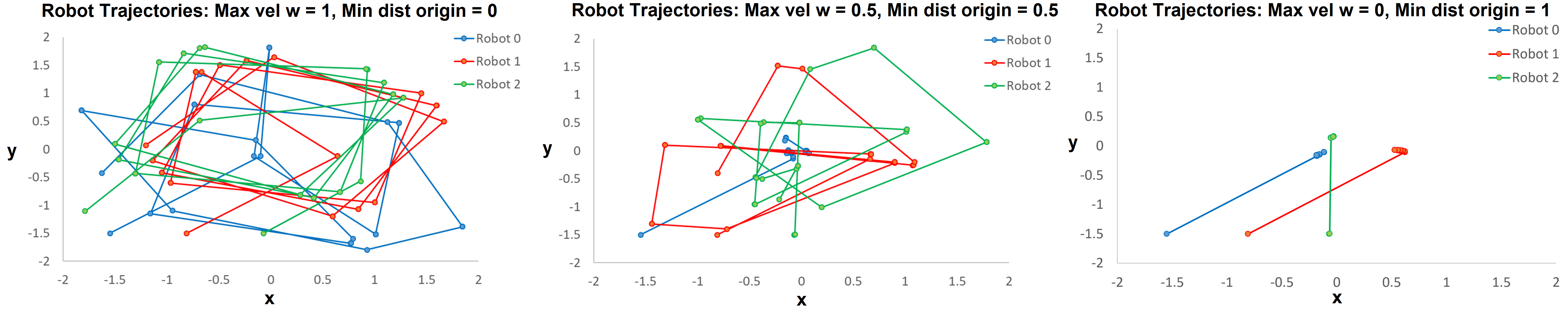}}
\caption{Low-fidelity simulator robot trajectories: Maximize velocity objective (left). Equal objective preference (middle). Minimize distance to origin (right).}
\label{fig:lowFidTraject1}
\end{figure*}

The evolved MO-NNs were capable of controlling the robot swarm in the desired manner in the low-fidelity python simulator. Figure \ref{fig:lowFidTraject1} illustrate the trajectories for the evolved MO-NN. 

The trajectories in Figure \ref{fig:lowFidTraject1} refer to the training objectives: Objective 1 - Maximize velocity and Objective 2 - Minimize distance to origin (center of arena). It is clear from Figure \ref{fig:lowFidTraject1} that the robot trajectories are significantly different when a high weighting is given to maximizing velocity compared to when a high weighting is given to minimizing distance to the origin. When a high weighting is given to maximizing velocity, the robots move in large circles around the arena. When a high weighting is given to minimizing distance to the origin, the robots move to the center of the arena and stay there. When an equal weighting is given to both objectives, the robots traverse the arena close to the center in smaller circles.

% The trajectories in Figure \ref{fig:lowFidTraject2} refer to the training objectives: Objective 1 - Maximize distance to origin and Objective 2 - Minimize distance to origin. Similarly, different behaviour can be observed for different objective weighting passed into the NN controller. When a high weight value for the maximize distance to the origin objective is selected, the robots remain close to the boundary. When the weight is then changed to minimize distance to the origin, the robots move to the center of the arena and stay there.

% \begin{figure}[h]
% \centerline{\includegraphics[width=0.9\textwidth]{Trajectories_MaxVelVSMinDistOrigin_2.png}}
% \caption{Low-fidelity simulator robot trajectories: Maximize velocity objective (top). Equal objective preference (bottom left). Minimize distance to origin (bottom right).}
% \label{fig:lowFidTraject1}
% \end{figure}

% \begin{figure}[h]
% \centerline{\includegraphics[width=1.0\textwidth]{Trajectories_MaxDistOriginVSMinDistOrigin.png}}
% \caption{Low-fidelity simulator robot trajectories: Maximize distance to origin (left). Equal objective preference (center). Minimize distance to origin (right).}
% \label{fig:lowFidTraject2}
% \end{figure}

% ****edit this***
% The convergence graph of the evolutionary multi-objective neural network is illustrated in Figure \ref{fig:convergence_pareto} (left). This figure illustrates that the algorithm has converged within the 20,000 evaluations of different robot neural network controller parameters.

\begin{figure}[h]
\centerline{\includegraphics[width=0.5\textwidth]{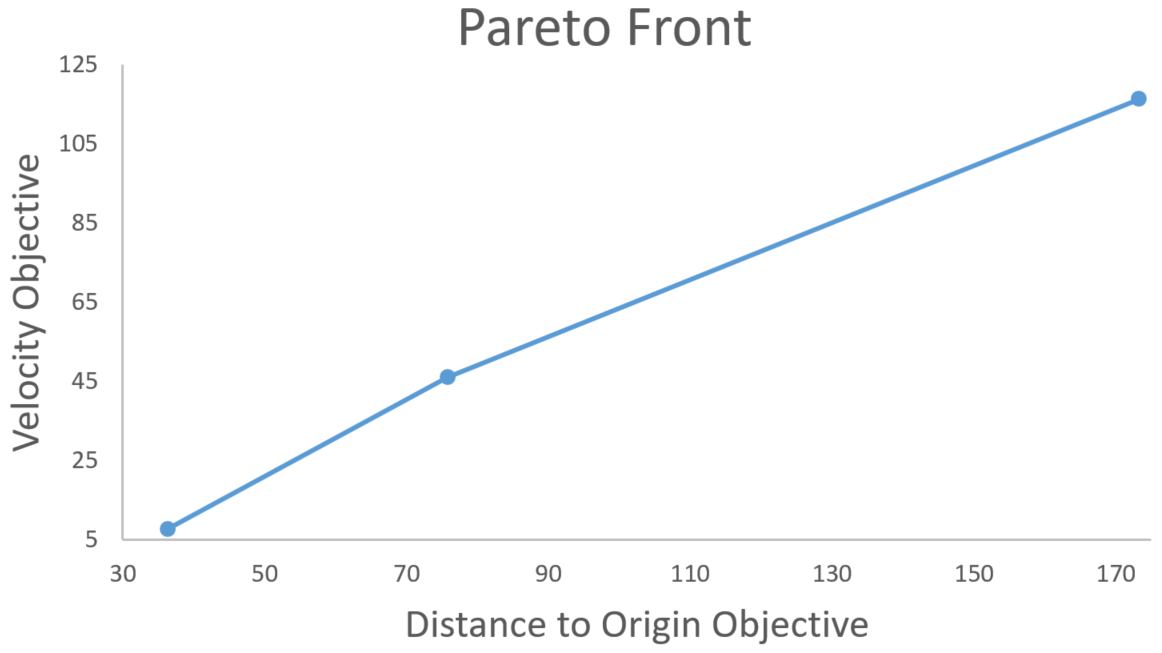}}
\caption{Pareto Front for evolving multi-objective neural network controller.}
\label{fig:convergence_pareto}
\end{figure}

% \begin{figure*}[h]
% \centerline{\includegraphics[width=0.95\textwidth]{Convergence_Pareto.png}}
% \caption{Convergence graph (left) and Pareto Front (right) for evolving multi-objective neural network controller.}
% \label{fig:convergence_pareto}
% \end{figure*}

% \begin{figure}[h]
% \centerline{\includegraphics[width=0.6\textwidth]{Convergence_MaxVel_MinDist.png}}
% \caption{Convergence graph for evolving multi-objective neural network controller.}
% \label{fig:convergence}
% \end{figure}

Figure \ref{fig:convergence_pareto} illustrates how the each of the objective function evaluations of the behaviour of the evolved neural network change as the objective weighting changes. When a maximum weighting is given to minimizing the distance to the origin objective, both the distance and velocity objective scores are lowest. Conversely when maximizing velocity, the distance to the origin also increases as the robots are moving with higher velocity around their environment. This graph illustrates how different behaviour can be observed from the swarm of robots using a single neural network by simply modifying the weighting for each objective function. No retraining is required for different objectives or different robots. This is the key advantage of the proposed MO-NN approach.

% \begin{figure}[h]
% \centerline{\includegraphics[width=0.6\textwidth]{ParetoFront.png}}
% \caption{Pareto Front displaying change in objective fitness by varying objective weighting.}
% \label{fig:ParetoFront}
% \end{figure}

\subsection{Testing Evolved Controllers in High-Fidelity Webots Simulator}

% \begin{figure*}[h]
% \centerline{\includegraphics[width=0.7\textwidth]{60SecondsOfSimulationTime.png}}
% \caption{Webots robot positions: (a) At the beginning of both simulations. (b) After 60 seconds of simulations where robots minimize distance to the origin. (c) After 60 seconds of simulations where robots maximize velocity.}
% \label{fig:Webots60Sec}
% \end{figure*}

After training in a low-fidelity Python simulator, the evolved MO-NN robot controller was then deployed to control robots implemented in the high-fidelity Webots simulator. The motivation for doing this was to test whether the behaviours of the evolved controllers have the potential to translate to simulators with more realistic physics without the need to adapt the motor commands.

Two simulations were conducted. In the first simulation, the robots' objective preference was to minimize distance to the origin, i.e., all robots were passed an objective weighting of 1 for the minimize distance to the origin objective, and 0 for the maximize velocity objective. In the second simulation, the robots' objective preference was to maximize velocity. It was observed that the trained MO-NNs exhibited similar behaviour when tested in the Webots environment. 

% Figure \ref{fig:Webots60Sec} illustrates the robots in their starting position at time $t=0$ seconds and also after approximately 60 seconds with max weighting for each objectives, i.e. minimizing distance to origin and maximizing velocity.

% \begin{figure}[h]
% \centerline{\includegraphics[width=0.8\textwidth]{60SecondsOfSimulationTime.png}}
% \caption{Webots robot positions: (a) At the beginning of both simulations. (b) After 60 seconds of simulations where robots minimize distance to the origin. (c) After 60 seconds of simulations where robots maximize velocity.}
% \label{fig:Webots60Sec}
% \end{figure}

When minimizing distance to the origin, the robots moved much slower. The robots gradually made their way to the center of the arena and do not adjust their position much thereafter. Each robot continually rotates in order to sense more of its environment. 

When maximizing velocity, the robots move continuously around the arena and do not stop at the center. There is a greater risk of collisions when moving with higher velocity. In order to reduce the severity of the collisions, the maximum wheel rotational speed was reduced from 80 radians/sec to 60 radians/sec. Without this speed reduction, robots collided with one another, overturned and remained immobilized. Note, this velocity clamping was applied in both simulations reported.

\subsection{Larger Robot Swarms in High-Fidelity Webots Simulator}

The next set of simulations conducted was to establish if the evolved MO-NN can scale to larger swarm sizes without retraining. The motivation for this was to test if the controllers evolved in an environment with fewer robots are robust enough to scale to environments with more robots and therefore greater chance of collisions. In order to test this, simulations were conducted for robot swarm sizes of 5 and 10 robots, using the evolved MO-NN trained in a swarm size of 3. 

It was observed that the robot swarm gather at the center of the arena after 60 seconds in all robot swarm sizes when minimizing distance to the origin. Different behaviour was observed when maximizing velocity. After 60 seconds, the robots are dispersed around the arena as they are traversing the arena with higher velocity.

%Figure \ref{fig:largerSwarms} depicts the positions of the robots after 60 seconds in each simulation conducted.

% ***maybe move this image to appendix***
% \begin{figure*}[h]
% \centerline{\includegraphics[width=0.8\textwidth]{RobotSwarmSize.png}}
% \caption{Robot positions after 60 seconds for 3 (left), 5 (centre), and 10 (right) robots when maximizing velocity (top) and minimizing distance to origin (bottom).}
% \label{fig:largerSwarms}
% \end{figure*}

% As Figure \ref{fig:largerSwarms} illustrates, 

% The evolved MO-NN scales well when deployed in larger robot swarms than it was trained in. From the bottom set of results in Figure \ref{fig:largerSwarms}, the robots gather at the center of the arena after 60 seconds in all robot swarm sizes when minimizing distance to the origin. The top set of results show different behaviour when maximizing velocity. The robots here are more dispersed as they are traversing the arena with higher velocity.

\begin{figure}[h!]
\centerline{\includegraphics[width=0.5\textwidth]{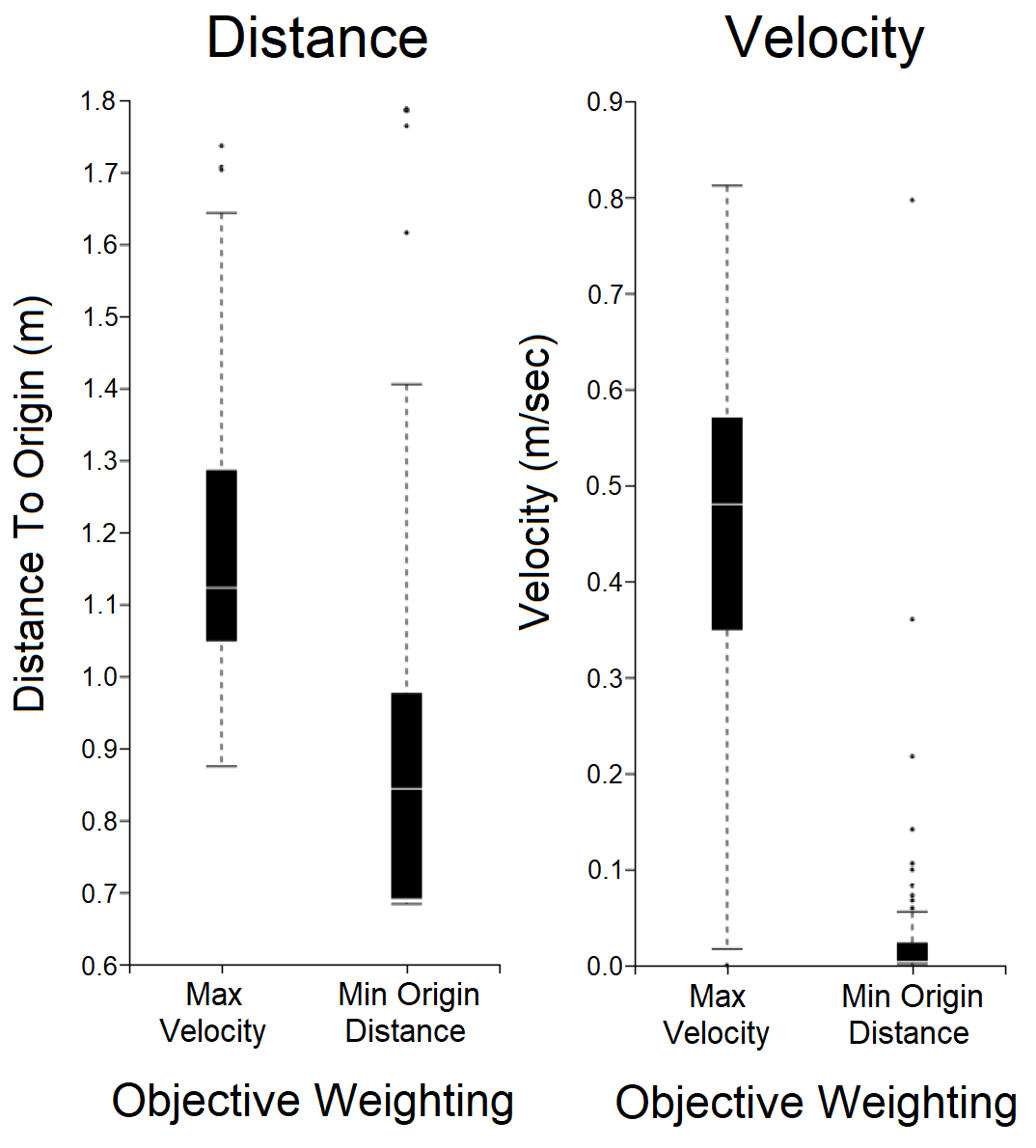}}
\caption{Distribution of distance to origin (left) and velocity (right) when varying objective weighting.}
\label{fig:boxPlot}
\end{figure}

Figure \ref{fig:boxPlot} illustrates the spread in the distance to the origin and the velocity averaged over 10 robots at each second in the 10 robot swarm for 60 seconds. This graph clearly illustrates how the robots travel with significantly higher velocity when the weighting on the velocity objective is at its maximum, compared to when a maximum weighting is applied to the distance to origin objective. This difference is statistically significant when compared using the two tailed Wilcoxon signed rank test, with significance level $\alpha = 1\%$. Similarly, when comparing the distance to the origin at each second, it can be seen that the distance is significantly lower when maximizing the weighting for the distance objective compared to maximizing the velocity objective, with a significance level $\alpha = 1\%$.

% \begin{figure}[h]
% \centerline{\includegraphics[width=0.5\textwidth]{boxPlots.png}}
% \caption{Distribution of distance to origin (left) and velocity (right) when varying objective weighting.}
% \label{fig:boxPlot}
% \end{figure}

% Figure \ref{fig:heatMap} presents a heatmap of the positions of all robots in the 10 robot swarm over 60 seconds when maximizing velocity (left) and minimizing distance to the origin (right). The figure on the left illustrates how the robots are more dispersed throughout the map when maximizing velocity. When minimizing distance to the origin, the robot positions are concentrated heavily in the center of the map. This is as expected.

\begin{figure}[h]
\centerline{\includegraphics[width=0.99\textwidth]{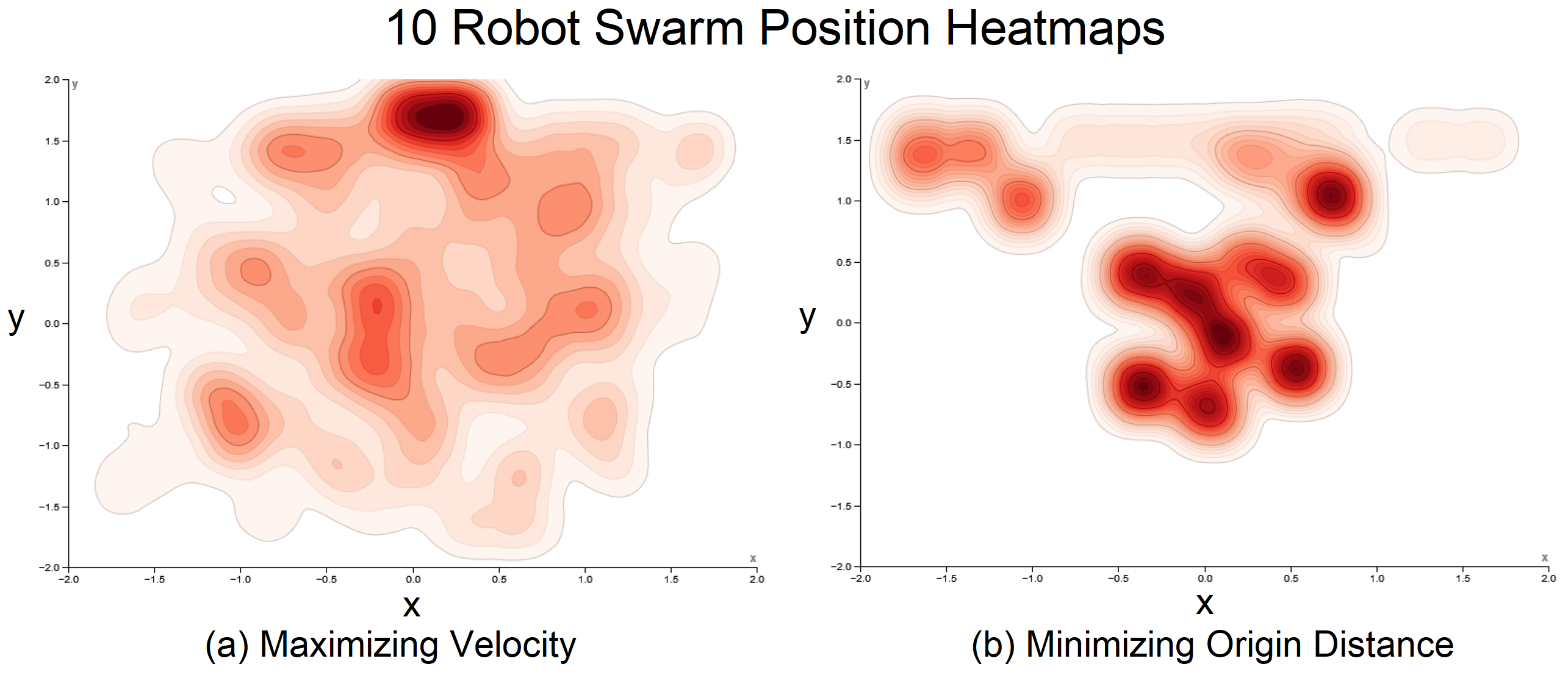}}
\caption{Position heatmaps of 10 robots over 60 seconds when maximizing velocity (a) and minimizing distance to origin (b).}
\label{fig:heatMap}
\end{figure}

Figure \ref{fig:heatMap} presents a heatmap of the positions of all robots in the 10 robot swarm over 60 seconds when maximizing velocity (a) and minimizing distance to the origin (b). The figure on the left illustrates how the robots are more dispersed throughout the map when maximizing velocity. When minimizing distance to the origin, the robot positions are concentrated heavily in the center of the map. This is as expected.

\begin{figure*}[h]
\centerline{\includegraphics[width=0.99\textwidth]{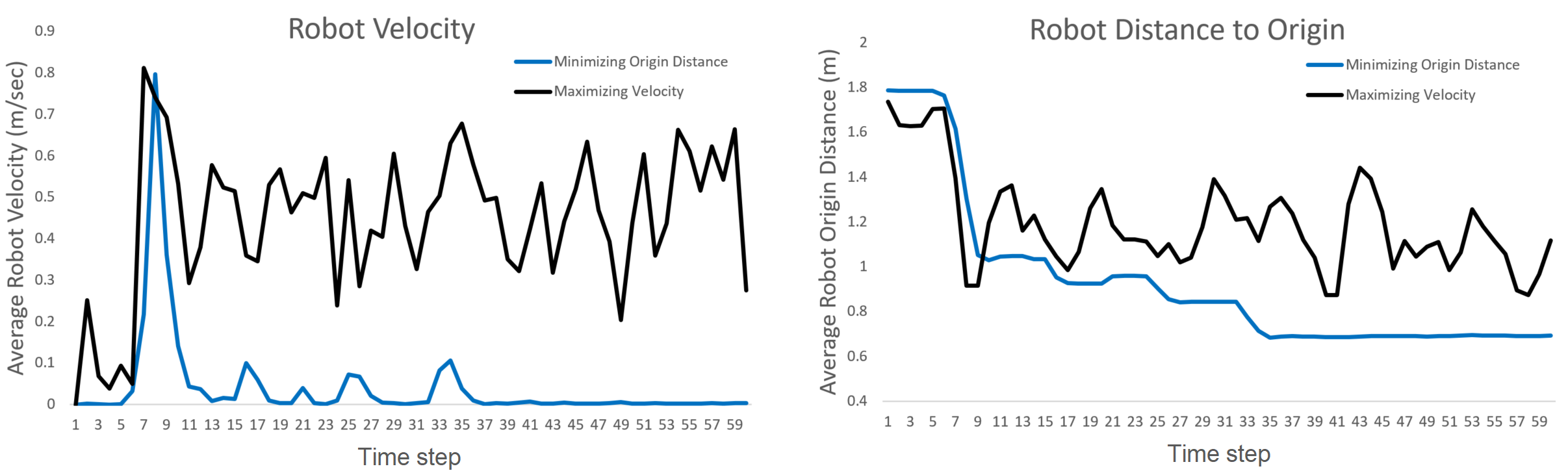}}
\caption{10 robot swarm average velocity (left) and distance to origin (right) over 60 seconds.}
\label{fig:metricsPerTime}
\end{figure*}

Figure \ref{fig:metricsPerTime} presents the robot velocity (left) and distance to origin (right) over 60 seconds of simulation time when maximizing the weighting to each objective. Under both objective preferences, the average velocity of the swarm is slow initially. When minimizing the distance to the origin, there is a large spike in velocity at time step 8. This corresponds in a significant reduction in the average distance to the origin. There are multiple smaller spikes in average velocity which further reduce the swarm distance to the origin. When maximizing velocity, there is a similar large increase in velocity early in the simulation. The average velocity of the robots then stabilize with some oscillations thereafter.

\section{Conclusion}

This research proposed an evolutionary multi-objective (MO) neural network (NN) for robot swarm control. The MO-NN was evolved using a low-fidelity Python simulator in an environment with 3 robots. The controller was then tested in a high-fidelity simulated environment developed using Webots. The MO-NN controller was then evaluated for a larger numbers of robots.

The primary findings of this research are:

1) It was demonstrated that the evolved MO-NN is an effective approach for controlling robots in a swarm in the presence of multiple objectives. As the weightings for each objectives were varied, the robots adjusted their behaviour to account for the preference for each objective. The velocities and distances to the origin during simulation were significantly different (p value $<1\%$) when comparing a maximum weighting for each objective. 

2) MO-NN evolved in a low-fidelity simulator can be transferred to a high-fidelity simulator and exhibit similar behaviour. This is because the low-fidelity simulator can replicate the relevant features of the high-fidelity simulator, specifically sensor input, decision making and updates to robot positions. This approach is therefore recommended to reduce the computational cost associated training NN controllers. This indicates that the evolved MO-NN controllers are robust to environmental changes.

3) The MO-NN evolved in a low-fidelity simulated environment with 3 robots scales well when tested in a high-fidelity simulated environment with 10 robots. No further NN training is required. This is a key advantage of the proposed evolutionary MO-NN approach as opposed to MO path planning.

There are multiple directions for future research that have stemmed from this research. In particular, the next natural step is to test the evolved MO-NN on a physical robot swarm outside of simulation, e.g. the DOTS swarm arena \cite{jones2022dots}. Another avenue for future research would be to investigate the scalability of the MO-NN to greater than two objectives. The experimental results presented in this paper is for a relatively small swarm size. It is hypothesized that the proposed approach would scale well for larger swarms. This will be evaluated in future work.

%%%%%%%%%%%%%%%%%%%%%%%%%%%%%%%%%%%%%%%%%%%%%%%%%%%%%%%%%%%%%%%%%%%%%%%%

%%% The acknowledgments section is defined using the "acks" environment
%%% (rather than an unnumbered section). The use of this environment 
%%% ensures the proper identification of the section in the article 
%%% metadata as well as the consistent spelling of the heading.

\section*{Acknowledgements}
%\begin{acks}
Research reported in this publication was funded by the Royal Irish Academy. The content of this publication is solely the responsibility of the authors and does not necessarily represent the official views of the Royal Irish Academy.

\begin{figure}[h]
\centerline{\includegraphics[width=0.4\textwidth]{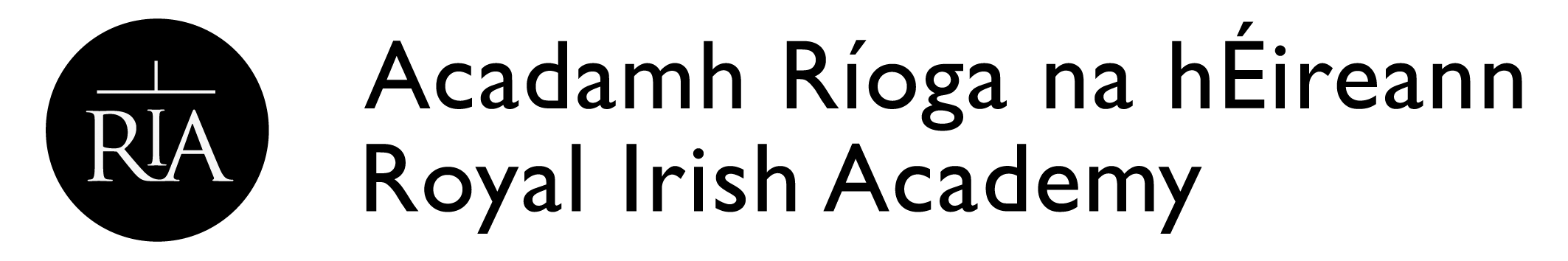}}
\label{fig:RIA}
\end{figure}
%\end{acks}

%%%%%%%%%%%%%%%%%%%%%%%%%%%%%%%%%%%%%%%%%%%%%%%%%%%%%%%%%%%%%%%%%%%%%%%%

%%% The next two lines define, first, the bibliography style to be 
%%% applied, and, second, the bibliography file to be used.

\bibliographystyle{ACM-Reference-Format} 
\bibliography{references}

\end{document}